\documentclass[runningheads]{llncs}
\usepackage[T1]{fontenc}
\usepackage{graphicx,verbatim,amsmath,multicol,multirow,amssymb}
\begin{document}
\title{Prototype-Enhanced Confidence Modeling for Cross-Modal Medical Image-Report Retrieval}
%
\author{Shreyank N Gowda\inst{1}\orcidID{0000-0002-4975-0705} \and
Xiaobo Jin\inst{2}\orcidID{0000-0003-1671-1379} \and
Christian Wagner\inst{1}\orcidID{0000-0002-6121-9722}}
\authorrunning{S. N. Gowda et al.}
%
\institute{School of Computer Science, The University of Nottingham, NG8 1BB
Nottingham, U.K. \and
Department of Intelligent Science, Xi’an Jiaotong-Liverpool University, China, 215123.
\email{shreyank.narayanagowda@nottingham.ac.uk}\\
}


\maketitle              
\begin{abstract}
In cross-modal retrieval tasks, such as image-to-report and report-to-image retrieval, accurately aligning medical images with relevant text reports is essential but challenging due to the inherent ambiguity and variability in medical data. Existing models often struggle to capture the nuanced, multi-level semantic relationships in radiology data, leading to unreliable retrieval results. To address these issues, we propose the Prototype-Enhanced Confidence Modeling (PECM) framework, which introduces multi-level prototypes for each modality to better capture semantic variability and enhance retrieval robustness. PECM employs a dual-stream confidence estimation that leverages prototype similarity distributions and an adaptive weighting mechanism to control the impact of high-uncertainty data on retrieval rankings. Applied to radiology image-report datasets, our method achieves significant improvements in retrieval precision and consistency, effectively handling data ambiguity and advancing reliability in complex clinical scenarios. We report results on multiple different datasets and tasks including fully supervised and zero-shot retrieval obtaining performance gains of up to 10.17\%, establishing in new state-of-the-art.

\keywords{Prototype  \and Uncertainty \and Retrieval \and Vision-Language.}

\end{abstract}
\section{Introduction}

Modern healthcare generates vast digital medical records including radiology images, creating both opportunity and challenge: effectively retrieving relevant cross-modal information is crucial for improving clinical decision-making. Cross-modal retrieval aligns and retrieves related data across modalities—using a query image to find relevant text or vice versa—bridging visual and textual information gaps. Adapting cross-modal retrieval to clinical domains introduces unique obstacles. Radiology images and reports contain specialized content with inherent ambiguities: overlapping visual features across conditions~\cite{li2020vispi} and limited descriptive granularity in reports~\cite{gowda2024masks}. Recent deep learning advances have improved retrieval capabilities through attention mechanisms and contrastive learning. MCR~\cite{mcr} creates robust image-report embeddings for medical retrieval tasks. Vision-Language Pre-training models like CLIP~\cite{radford2021learning} and its medical adaptations~\cite{biomedclip} enhance performance by transferring knowledge to downstream medical tasks. Other works~\cite{gowda2024masks,convirt,maskclip,cxrclip,gowda2025distribution} further align visual and linguistic medical representations. Most existing methods assume consistent data quality~\cite{maskclip,cxrclip,mcr}, rarely true in medical imaging. Prototype-based methods~\cite{li2024prototype} capture diverse semantic structures by learning representative prototypes for each modality. Whilst uncertainty has been explored in a wide variety of contexts~\cite{ghesu2021quantifying,yang2021uncertainty,gowda2024cc,baumgartner2019phiseg}, Li et al.~\cite{li2024prototype} are the only ones to quantify aleatoric uncertainty in cross-modal retrieval, improving reliability by accounting for data quality variations.

We propose a novel \textit{Prototype-Enhanced Confidence Modeling (PECM)} framework to improve robustness of medical cross-modal retrieval by quantifying representation uncertainty. Unlike traditional models, PECM explicitly considers ambiguity in radiology data through multi-level prototypes that capture different semantic aspects within images and reports. This enables representation of subtle differences across cases, improving retrieval accuracy where conventional models might be misled by ambiguous features. While standard cross-modal retrieval embeds images and text into a shared space using cosine similarity, PECM introduces a \textit{dual-stream confidence estimation} that models uncertainty by assessing dissimilarity within prototype similarity distributions and employs an \textit{adaptive weighting mechanism}. This adjusts the influence of high-uncertainty data points, balancing accuracy and reliability without disregarding informative cases. The resulting re-ranking system prioritizes high-confidence matches while still considering lower-confidence data, increasing robustness against ambiguity.

\noindent Our key contributions are:
\newline \noindent \textbf{1. Prototype-Enhanced Representation:} Multi-level prototypes represent unique semantic categories within each modality, providing finer-grained understanding of radiology data by capturing variations at different semantic levels, enhancing generalization across diverse medical datasets.
\newline \noindent \textbf{2. Dual-Stream Confidence Estimation:} Using dissimilarity within prototype distributions as an uncertainty proxy, our model adaptively weights uncertain cases. This achieves informed balancing of information, effectively balancing accuracy and informativeness without discarding useful data.
\newline \noindent \textbf{3. Adaptive Re-ranking Mechanism:} We implement adaptive ranking that leverages confidence information to dynamically adjust retrieval order, prioritizing high-confidence matches. This enriches retrieval by integrating uncertainty as an informative signal, enhancing robustness.

Through experiments on multiple medical retrieval tasks, PECM significantly outperforms traditional models in accuracy and consistency, demonstrating strong potential for clinical applications.

\section{Method}

\begin{figure*}[ht]
    \centering
    \includegraphics[width=0.95\textwidth]{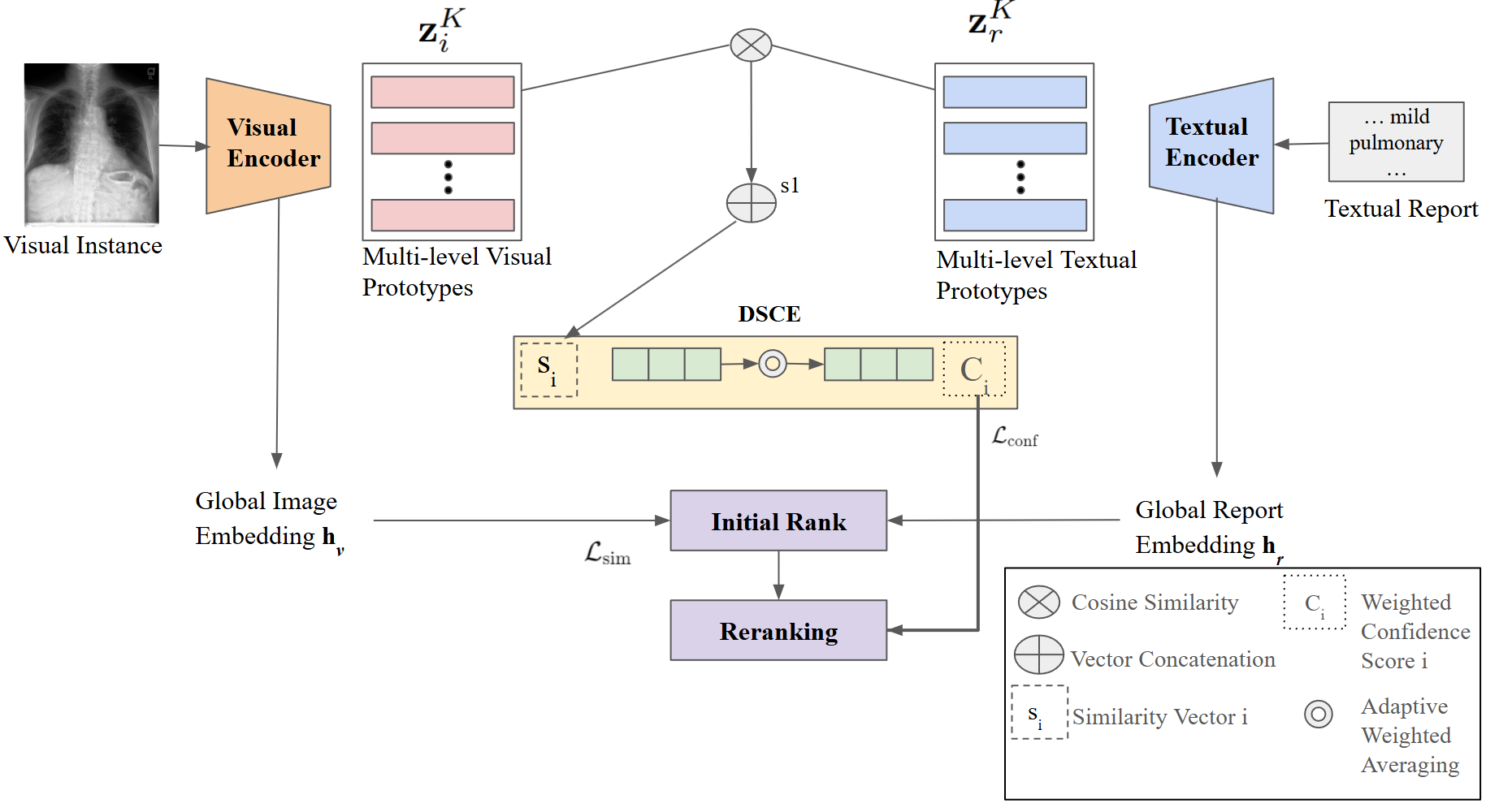}
    \caption{
        PECM processes paired image-report data through encoders to generate multi-level prototypes. The Dual-Stream Confidence Estimation calculates similarities between corresponding prototypes, weighted to produce confidence scores. Initial ranking uses global similarity, while re-ranking incorporates confidence-weighted similarity. The model is trained with a combination of similarity loss (\( \mathcal{L}_{\text{sim}} \)), confidence loss (\( \mathcal{L}_{\text{conf}} \)), and diversity loss (\( \mathcal{L}_{\text{diversity}} \)) (not illustrated here for simplicity).
    }
    \label{fig:overview}
\end{figure*}

This section introduces the Prototype-Enhanced Confidence Modeling (PECM) framework for cross-modal retrieval, designed to quantify uncertainty in medical imaging data. PECM consists of three key components: multi-level prototype construction, dual-stream confidence estimation, and adaptive re-ranking. An overview appears in Figure~\ref{fig:overview}.

\subsection{Multi-Level Prototype Construction}

\begin{figure}[h!]
    \centering
    \includegraphics[width=0.75\columnwidth]{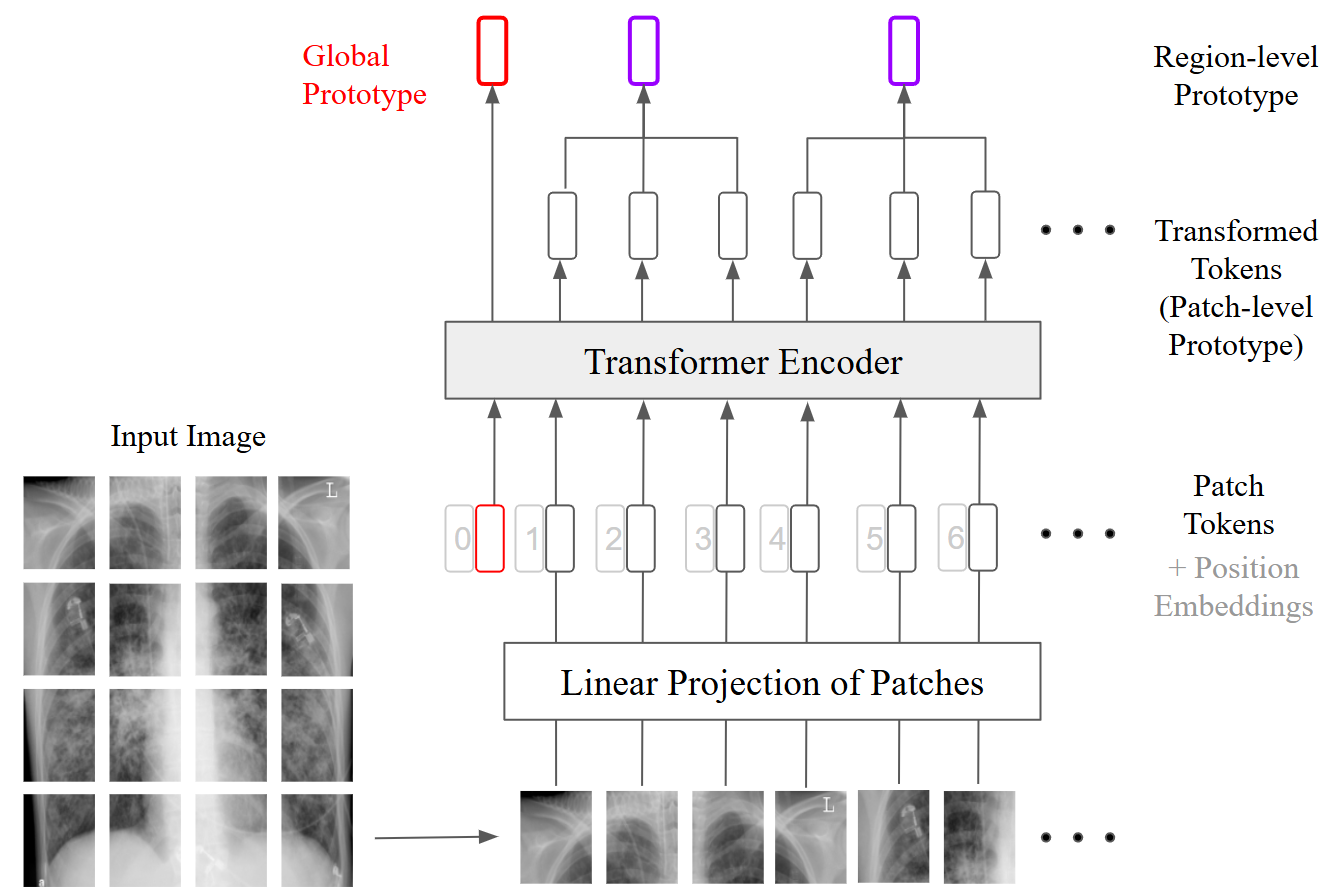}
    \caption{
        \textbf{Multi-Level Prototype Construction.} Illustration of prototype calculation: images are patched, processed through transformer encoding, averaged into patch-level prototypes, grouped into regional prototypes, and combined with global prototype to form multi-level representation. 16 patches for illustration only.
    }
    \label{fig:prototype}
\end{figure}

To address variability in medical data, we create multi-level prototypes for each modality to capture diverse semantic meanings. With $K$ prototypes per modality, each represents a distinct level of semantic granularity.

For images, our visual encoder (ViT-B) divides 224×224 images into 16×16 patches, producing 196 patches (14×14 grid). To reduce computation, we group these into 3×3 grids, creating 25 regional prototypes across the image (5×5 regions). We add a global prototype from the [CLS] token, resulting in $K=26$ prototypes per image (Figure~\ref{fig:prototype}). For text, we divide each report into $K-1$ sentence groups and process each through a paragraph embedding model~\cite{doc2vec} to obtain region-level prototypes. These combine with the global document prototype to form $K$ prototypes per report. We denote image and report prototypes as $\mathbf{Z}_i = \{ \mathbf{z}_i^1, \mathbf{z}_i^2, \dots, \mathbf{z}_i^K \}$ and $\mathbf{Z}_r = \{ \mathbf{z}_r^1, \mathbf{z}_r^2, \dots, \mathbf{z}_r^K \}$, respectively.

\subsection{Dual-Stream Confidence Estimation}

Our dual-stream confidence estimation quantifies image-text match reliability by aligning corresponding prototypes. Image and report modalities are represented as prototype sets $\mathbf{Z}_i = \{ \mathbf{z}_i^1, \mathbf{z}_i^2, \dots, \mathbf{z}_i^K \}$ and $\mathbf{Z}_r = \{ \mathbf{z}_r^1, \mathbf{z}_r^2, \dots, \mathbf{z}_r^K \}$. For each prototype pair $(\mathbf{z}_i^k, \mathbf{z}_r^k)$, we compute cosine similarity:

\begin{equation}
   \text{sim}(\mathbf{z}_i^k, \mathbf{z}_r^k) = \frac{\mathbf{z}_i^k \cdot \mathbf{z}_r^k}{\|\mathbf{z}_i^k\| \|\mathbf{z}_r^k\|}
\end{equation}

These similarities form vector $\mathbf{s} = [\text{sim}(\mathbf{z}_i^1, \mathbf{z}_r^1), \text{sim}(\mathbf{z}_i^2, \mathbf{z}_r^2), \dots, \text{sim}(\mathbf{z}_i^K, \mathbf{z}_r^K)]$. The final confidence score $c$ is their weighted average:

\begin{equation}
   c = \frac{1}{K} \sum_{k=1}^{K} \text{sim}(\mathbf{z}_i^k, \mathbf{z}_r^k) \cdot w_k
\end{equation}

where adaptive weights $w_k$ emphasize high-confidence pairs and reduce influence of ambiguous matches.

\subsection{Adaptive Re-ranking Mechanism}

PECM initially ranks retrieval based on global similarity between image and report representations:

\begin{equation}
   \text{sim}_{\text{initial}}(v, t) = \frac{\mathbf{h}_v \cdot \mathbf{h}_t}{\|\mathbf{h}_v\| \|\mathbf{h}_t\|}
\end{equation}

where global embeddings $\mathbf{h}_v$ and $\mathbf{h}_t$ are weighted averages of their respective prototypes:

\begin{equation}
   \mathbf{h}_v = \sum_{k=1}^{K} w_k \mathbf{z}_i^k, \quad \mathbf{h}_t = \sum_{k=1}^{K} w_k \mathbf{z}_r^k
\end{equation}

with shared weights $w_k$ for cross-modal alignment. For query image $v$ and candidate reports $\{t_j\}_{j=1}^N$, initial ranking is:

\begin{equation}
   \text{rank}_{\text{initial}}(v, t_j) = \text{Rank}_{\text{desc}}\left(\text{sim}_{\text{initial}}(v, t_j) : j = 1, \dots, N \right)
\end{equation}

We then calculate confidence scores for adaptive re-ranking:

\begin{equation}
   C(v, t_j) = \frac{1}{K} \sum_{k=1}^{K} \text{sim}(\mathbf{z}_i^k, \mathbf{z}_r^k) \cdot w_k
\end{equation}

The final score weights initial similarity with confidence:

\begin{equation}
   R(v, t_j) = \text{sim}_{\text{initial}}(v, t_j) \cdot C(v, t_j)
\end{equation}

Final ranking prioritizes high-confidence matches:

\begin{equation}
   \text{rank}_{\text{final}}(v, t_j) = \text{Rank}_{\text{desc}}\left(R(v, t_j) : j = 1, \dots, N \right)
\end{equation}

\subsection{Objective Function}

PECM training uses a triple-component objective function:

\begin{equation}
    \mathcal{L} = \mathcal{L}_{\text{sim}} + \lambda \mathcal{L}_{\text{conf}} + \mu \mathcal{L}_{\text{div}}
\end{equation}

where $\lambda$ and $\mu$ are balancing hyperparameters. The similarity loss uses contrastive learning to promote match quality:

\begin{equation}
    \mathcal{L}_{\text{sim}} = -\log \frac{\exp(\text{sim}_{\text{initial}}(v, t))}{\sum_{j=1}^{N} \exp(\text{sim}_{\text{initial}}(v, t_j))}
\end{equation}

The confidence loss optimizes prototype weights, penalizing low-confidence matches:

\begin{equation}
    \mathcal{L}_{\text{conf}} = \frac{1}{N} \sum_{i=1}^{N} \left(1 - C(v_i, t_i)\right)^2
\end{equation}

where $C(v_i, t_i)$ represents weighted similarity between corresponding prototypes. The diversity loss ensures prototypes within each modality remain distinct:

\begin{equation}
    \mathcal{L}_{\text{div}} = \sum_{k=1}^{K} \sum_{l \neq k}^{K} \left(1 - \text{sim}(\mathbf{z}^k, \mathbf{z}^l)\right)^2
\end{equation}

where $\mathbf{z}^k$ and $\mathbf{z}^l$ are prototypes within the same modality, encouraging each to capture unique semantic features.

This objective balances similarity, confidence, and diversity for accurate and reliable retrieval across ambiguous medical data. We set $\mu$ and $\lambda$ as 1 based on experimental results.

\section{Experimental Analysis}

\subsection{Datasets}

Following MCR~\cite{mcr}, we use MIMIC-CXR~\cite{mimic} (377,100 images, 65,379 patients, 227,835 reports) for cross-modal retrieval, extracting 'findings' sections and excluding incomplete samples. For generalizable retrieval, we combine RadImageNet~\cite{radimagenet}, NIH14~\cite{nih}, MIMIC~\cite{mimic}, and CheXpert~\cite{chexpert} per~\cite{foundation}, creating a diverse dataset (1.6M+ images, 4 modalities, 12 anatomical regions, 185 classes) with duplicate removal, label standardization, and single-label focus. Additional evaluations use ROCO from ImageCLEF 2023~\cite{roco} (60,918/10,437/10,473 train/val/test images with captions and CUIs) and MURA~\cite{mura} (40,895 musculoskeletal radiographs across seven body parts).

\subsection{Comparison to State-of-the-art}

\subsubsection{Cross-modal Retrieval}

\vspace{-5mm}

\begin{table*}[ht]
\centering
\caption{The performance comparison of different cross-modal retrieval methods on the MIMIC-CXR dataset. VLP and MVLP represent Vision-Language Pre-training and Medical Vision-Language Pre-training, respectively. MMU and CMA represent Multi-Modal Fusion and Cross-Modal Alignment, respectively.}
\resizebox{0.98\textwidth}{!}{
\begin{tabular}{|l|c|ccc|ccc|}
\hline
\multirow{2}{*}{Method} & \multirow{2}{*}{Category} & \multicolumn{3}{c|}{I $\rightarrow$ R} & \multicolumn{3}{c|}{R $\rightarrow$ I} \\
                        &                           & Recall@1 & Recall@5 & Recall@10           & Recall@1 & Recall@5 & Recall@10 \\
\hline
ConVIRT~\cite{convirt}                 & MVLP, CMA                 & 6.765\%  & 22.421\% & 32.219\%            & 6.647\%  & 16.151\% & 23.832\% \\
REFERS~\cite{refers}                  & MVLP, CMA                 & 8.294\%  & 25.765\% & 36.910\%            & 8.607\%  & 18.929\% & 28.755\% \\
LoVT~\cite{lovt}                    & MVLP, CMA                 & 10.601\% & 30.715\% & 42.017\%            & 11.504\% & 24.855\% & 36.067\% \\
DiVE~\cite{lovt}                   & CMA                       & 12.960\% & 35.070\% & 48.419\%            & 13.294\% & 26.993\% & 38.780\% \\
CLIP~\cite{radford2021learning}                    & VLP, CMA                  & 14.256\% & 36.392\% & 48.263\%            & 14.700\% & 29.342\% & 40.160\% \\
BLIP2~\cite{blip}                   & VLP, CMA                  & 17.159\% & 41.498\% & 53.447\%            & 18.875\% & 34.974\% & 47.657\% \\
M3AE~\cite{m3ae}                    & MVLP, MMU                 & 15.267\% & 42.898\% & 56.739\%            & 17.171\% & 36.153\% & 50.135\% \\
MaskCLIP~\cite{maskclip}                & MVLP, CMA                 & 18.170\% & 43.339\% & 55.962\%            & 19.301\% & 36.531\% & 49.058\% \\
CXR-CLIP~\cite{cxrclip}                & MVLP, CMA                 & 20.736\% & 45.853\% & 58.968\%            & 22.985\% & 40.477\% & 52.484\% \\
MCR + MbA~\cite{mcr}              & MVLP, CMA                 & 24.598\% & 52.281\% & 65.241\%            & 27.354\% & 45.709\% & 58.758\% \\
SENSE~\cite{sense} & MVLP, CMA & 19.500\% & 45.100\% & 57.300\% & 19.800\% & 46.200\% & 57.500\% \\
\hline
\textbf{PECM (Ours)} & MVLP, CMA & \textbf{28.871\%} & \textbf{58.644\%} & \textbf{69.691\%} & \textbf{32.431\%} & \textbf{50.946\%} & \textbf{64.252\%} \\
\hline
\end{tabular}
}
\label{tab:cross-modal}
\end{table*}

For cross-modal retrieval, we evaluate image-to-report (I$\rightarrow$R) and report-to-image (R$\rightarrow$I) tasks. Following MCR~\cite{mcr}, we use Recall@K (K $\in$ \{1,5,10\}) as our metric, measuring the percentage of correct matches within the top K retrievals. For R$\rightarrow$I, recall is capped by min(K, number of images linked to the report). 
Table~\ref{tab:cross-modal} shows our approach achieves performance gains up to 6.363\% on MIMIC-CXR.

\subsubsection{Generalizable Retrieval}

\begin{table*}[ht]
\centering
\caption{Content-Based Image Retrieval performance of the evaluated models on the combined dataset.}
\begin{tabular}{|l|cccc|cccc|}
\hline
\multirow{2}{*}{Method} & \multicolumn{4}{c|}{Sample-wise (micro)} & \multicolumn{4}{c|}{Class-wise (macro)} \\
                        & P@1  & P@3  & P@5  & P@10              & P@1  & P@3  & P@5  & P@10 \\
\hline
ViT~\cite{vit}                    & 0.560 & 0.554 & 0.550 & 0.543           & 0.217 & 0.204 & 0.199 & 0.190 \\
SAM~\cite{sam}                     & 0.520 & 0.517 & 0.515 & 0.511           & 0.197 & 0.187 & 0.182 & 0.175 \\
MedSAM~\cite{medsam}                  & 0.489 & 0.483 & 0.478 & 0.470           & 0.154 & 0.146 & 0.142 & 0.134 \\
CLIP~\cite{radford2021learning}                    & 0.571 & 0.567 & 0.565 & 0.559           & 0.222 & 0.213 & 0.209 & 0.202 \\
MedCLIP~\cite{medclip}                 & 0.566 & 0.561 & 0.559 & 0.554           & 0.223 & 0.211 & 0.205 & 0.198 \\
BiomedCLIP~\cite{biomedclip}              & 0.594 & 0.590 & 0.588 & 0.583           & 0.240 & 0.230 & 0.224 & 0.217 \\
M3AE~\cite{m3ae}                     & 0.513 & 0.507 & 0.503 & 0.497           & 0.183 & 0.177 & 0.172 & 0.166 \\
DINOv2~\cite{dino}                  & 0.553 & 0.548 & 0.545 & 0.540           & 0.219 & 0.204 & 0.199 & 0.192 \\
X-MIR~\cite{x-mir}                   & 0.478 & 0.471 & 0.466 & 0.457           & 0.149 & 0.142 & 0.138 & 0.131 \\
KL-CVR~\cite{kl-cvr}              & 0.602 & 0.596 & 0.593 & 0.585           & 0.245 & 0.234 & 0.229 & 0.211 \\
\hline
\textbf{PECM (Ours)} & \textbf{0.631} & \textbf{0.627} & \textbf{0.616} & \textbf{0.605} & \textbf{0.257} & \textbf{0.249} & \textbf{0.241} & \textbf{0.219} \\
\hline
\end{tabular}
\label{tab:cbir-combined}
\end{table*}

Using our pre-trained visual encoder, we compare against ViT~\cite{vit}, SAM~\cite{sam}, MedSAM~\cite{medsam}, CLIP~\cite{radford2021learning}, MedCLIP~\cite{medclip}, BiomedCLIP~\cite{biomedclip}, M3AE~\cite{m3ae}, DINOv2~\cite{dino}, X-MIR~\cite{x-mir}, and KL-CVR~\cite{kl-cvr}. We report both micro- and macro-averaged scores to ensure fair evaluation across imbalanced datasets. Table~\ref{tab:cbir-combined} shows up to 2.9\% improvement.

\subsubsection{Zero-Shot Retrieval}

\vspace{-9mm}

\begin{table*}[ht]
\centering
\caption{Generalizing CIBR to other datasets}
\begin{tabular}{|l|ccc|ccc|}
\hline
\multirow{2}{*}{Methods} & \multicolumn{3}{c|}{ROCO CUI@K} & \multicolumn{3}{c|}{MURA P@K} \\
                         & @5       & @10      & @50       & @5       & @10      & @30       \\
\hline
\multicolumn{7}{|l|}{\textit{General initialization methods}} \\
\hline

ImageNet                 & 35.53    & 37.39    & 39.18     & 84.42    & 77.23    & 70.04     \\
CLIP~\cite{radford2021learning}                     & 35.72    & 37.39    & 39.20     & 65.70    & 54.38    & 46.55     \\
\hline
\multicolumn{7}{|l|}{\textit{In-domain initialization methods: Fully supervised}} \\
\hline
CLIP~\cite{radford2021learning}                     & 39.85    & 40.84    & 44.86     & 69.96    & 57.82    & 50.55     \\
BioMed CLIP~\cite{biomedclip}              & 42.49    & 45.07    & \textbf{47.23}     & 74.34    & 65.89    & 56.50     \\
KL-CVR~\cite{kl-cvr}            & 40.68    & 43.44    & 46.46     & \textbf{77.83}    & 68.56    & 60.80     \\
\hline
\multicolumn{7}{|l|}{\textit{In-domain initialization methods: Zero-Shot}}\\
\hline
KL-CVR~\cite{kl-cvr} & 33.67 & 35.51 & 38.72 & 66.52 & 64.65 & 55.89 \\
\textbf{PECM (Ours)} & \textbf{42.95} & \textbf{45.23} & \textbf{47.23} & 76.69 & \textbf{68.59} & \textbf{60.92} \\

\hline
\end{tabular}
\label{tab:zero-shot}
\end{table*}

We compare PECM against state-of-the-art methods on ROCO and MURA without fine-tuning. Our model, pre-trained across various modalities and anatomical regions, outperforms methods specifically fine-tuned on these benchmarks. While further fine-tuning yields slight improvements, PECM achieves competitive or superior performance in zero-shot settings, surpassing current state-of-the-art~\cite{kl-cvr}. This demonstrates PECM's robust generalization across diverse medical datasets, with performance gains up to 10.17\% as shown in Table~\ref{tab:zero-shot}. 

\subsection{Ablation Study}

\begin{table}[ht]
\begin{minipage}[t]{0.48\textwidth}
\centering
\caption{Ablation study of the proposed components. PER = Prototype-Enhanced Representation, DSCE = Dual-Stream Confidence Estimation, ARR = Adaptive Re-ranking. Reported results are for Recall @ 5 on MIMIC-CXR.}
\begin{tabular}{|c|c|c|c|c|}
\hline
PER & DSCE & ARR & I $\rightarrow$ R & R $\rightarrow$ I \\
\hline
\texttimes & \texttimes & \texttimes & 36.572\% & 29.781\% \\
\checkmark & \texttimes & \texttimes & 44.392\% & 37.751\% \\
\texttimes & \checkmark & \texttimes & 41.922\% & 36.345\% \\
\texttimes & \texttimes & \checkmark & 39.472\% & 33.915\% \\
\checkmark & \checkmark & \texttimes & 54.452\% & 45.278\% \\
\checkmark & \texttimes & \checkmark & 50.794\% & 45.046\% \\
\texttimes & \checkmark & \checkmark & 52.304\% & 47.116\% \\
\checkmark & \checkmark & \checkmark & \textbf{58.644}\% & \textbf{50.946}\% \\
\hline
\end{tabular}
\label{tab:ablation}
\end{minipage}
\hfill
\begin{minipage}[t]{0.48\textwidth}
\centering
\caption{Ablation study of individual loss components. Performance is reported as Recall @ 5 for image-to-report (I$\rightarrow$R) and report-to-image (R$\rightarrow$I) tasks. Results on MIMIC-CXR Dataset.}
\begin{tabular}{|c|c|c|c|c|}
\hline
$\mathcal{L}_{\text{sim}}$ & $\mathcal{L}_{\text{conf}}$ & $\mathcal{L}_{\text{div}}$ & I$\rightarrow$R & R$\rightarrow$I \\
\hline
\checkmark & \texttimes & \texttimes & 55.2 & 46.8 \\
\texttimes & \checkmark & \texttimes & 50.1 & 43.5 \\
\texttimes & \texttimes & \checkmark & 48.5 & 42.0 \\
\checkmark & \checkmark & \texttimes & 57.1 & 49.0 \\
\checkmark & \texttimes & \checkmark & 56.8 & 48.1 \\
\texttimes & \checkmark & \checkmark & 52.7 & 44.8 \\
\checkmark & \checkmark & \checkmark & \textbf{58.6} & \textbf{50.9} \\
\hline
\end{tabular}
\label{tab:loss-ablation}
\end{minipage}
\end{table}

We assess each component's contribution through ablation studies of Prototype-Enhanced Representation (PER), Dual-Stream Confidence Estimation (DSCE), and Adaptive Re-ranking (ARR). Table~\ref{tab:ablation} shows Recall@5 for I$\rightarrow$R and R$\rightarrow$I tasks. The baseline without these components performs poorest. Adding PER improves performance by capturing semantic nuances across modalities. DSCE further boosts recall through enhanced match reliability. The full PECM framework with all components achieves the best results, demonstrating their complementary effects. We also ablate the different loss functions in Table~\ref{tab:loss-ablation}.

\subsection{Implementation Details}

PECM is implemented in PyTorch using two pre-trained encoders: ViT-B (visual) and BERT (text). For Prototype-Enhanced Representation, images are divided into 16 patches, processed through ViT, and grouped into 3×3 regions to create 25 regional prototypes plus one global [CLS] prototype. Text prototypes are generated from doc2vec paragraph embeddings. DSCE computes prototype similarities with adaptive weights learned during training. Initial ranking uses global similarity, while Adaptive Re-ranking incorporates confidence scores to prioritize reliable matches. Training uses image-report pairs from multiple medical datasets with three loss components (contrastive, confidence, diversity) for 30 epochs (batch size 32, Adam optimizer, cosine annealing from 1e-4).

\section{Conclusion}

In this work, we addressed cross-modal retrieval challenges in medical imaging by developing the Prototype-Enhanced Confidence Modeling (PECM) framework. PECM introduces multi-level prototypes for both image and text modalities, capturing fine-grained semantic layers like anatomical and pathological details. Our Dual-Stream Confidence Estimation quantifies prototype alignment reliability, while Adaptive Re-ranking prioritizes confident matches. Experiments demonstrate PECM's superior performance across multiple medical retrieval tasks. By effectively handling both semantic complexity and uncertainty, PECM advances the state-of-the-art in medical cross-modal retrieval, providing a robust solution with clear potential for clinical applications.

\bibliographystyle{splncs04}
\bibliography{main}

\begin{thebibliography}{10}
\providecommand{\url}[1]{\texttt{#1}}
\providecommand{\urlprefix}{URL }
\providecommand{\doi}[1]{https://doi.org/#1}

\bibitem{baumgartner2019phiseg}
Baumgartner, C.F., Tezcan, K.C., Chaitanya, K., H{\"o}tker, A.M., Muehlematter, U.J., Schawkat, K., Becker, A.S., Donati, O., Konukoglu, E.: Phiseg: Capturing uncertainty in medical image segmentation. In: International Conference on Medical Image Computing and Computer-Assisted Intervention. pp. 119--127. Springer (2019)

\bibitem{m3ae}
Chen, Z., Du, Y., Hu, J., Liu, Y., Li, G., Wan, X., Chang, T.H.: Multi-modal masked autoencoders for medical vision-and-language pre-training. In: International Conference on Medical Image Computing and Computer-Assisted Intervention. pp. 679--689. Springer (2022)

\bibitem{foundation}
Denner, S., Zimmerer, D., Bounias, D., Bujotzek, M., Xiao, S., Kausch, L., Schader, P., Penzkofer, T., J{\"a}ger, P.F., Maier-Hein, K.: Leveraging foundation models for content-based medical image retrieval in radiology. arXiv preprint arXiv:2403.06567  (2024)

\bibitem{maskclip}
Dong, X., Bao, J., Zheng, Y., Zhang, T., Chen, D., Yang, H., Zeng, M., Zhang, W., Yuan, L., Chen, D., et~al.: Maskclip: Masked self-distillation advances contrastive language-image pretraining. In: Proceedings of the IEEE/CVF Conference on Computer Vision and Pattern Recognition. pp. 10995--11005 (2023)

\bibitem{vit}
Dosovitskiy, A., Beyer, L., Kolesnikov, A., Weissenborn, D., Zhai, X., Unterthiner, T., Dehghani, M., Minderer, M., Heigold, G., Gelly, S., et~al.: An image is worth 16x16 words. arXiv preprint arXiv:2010.11929  \textbf{7} (2020)

\bibitem{ghesu2021quantifying}
Ghesu, F.C., Georgescu, B., Mansoor, A., Yoo, Y., Gibson, E., Vishwanath, R., Balachandran, A., Balter, J.M., Cao, Y., Singh, R., et~al.: Quantifying and leveraging predictive uncertainty for medical image assessment. Medical Image Analysis  \textbf{68},  101855 (2021)

\bibitem{gowda2024cc}
Gowda, S.N., Clifton, D.A.: Cc-sam: Sam with cross-feature attention and context for ultrasound image segmentation. In: European Conference on Computer Vision. pp. 108--124. Springer (2024)

\bibitem{gowda2024masks}
Gowda, S.N., Clifton, D.A.: Masks and manuscripts: Advancing medical pre-training with end-to-end masking and narrative structuring. In: International Conference on Medical Image Computing and Computer-Assisted Intervention. pp. 426--436. Springer (2024)

\bibitem{gowda2025distribution}
Gowda, S.N., Zhang, R., Gu, X., Weng, Y., Yang, L.: Distribution-based masked medical vision-language model using structured reports. arXiv preprint arXiv:2507.21794  (2025)

\bibitem{x-mir}
Hu, B., Vasu, B., Hoogs, A.: X-mir: Explainable medical image retrieval. In: Proceedings of the IEEE/CVF Winter Conference on Applications of Computer Vision. pp. 440--450 (2022)

\bibitem{roco}
Ionescu, B., M{\"u}ller, H., Dr{\u{a}}gulinescu, A.M., Yim, W.W., Ben~Abacha, A., Snider, N., Adams, G., Yetisgen, M., R{\"u}ckert, J., Garc{\'\i}a Seco~de Herrera, A., et~al.: Overview of the imageclef 2023: Multimedia retrieval in medical, social media and internet applications. In: International Conference of the Cross-Language Evaluation Forum for European Languages. pp. 370--396. Springer (2023)

\bibitem{chexpert}
Irvin, J., Rajpurkar, P., Ko, M., Yu, Y., Ciurea-Ilcus, S., Chute, C., Marklund, H., Haghgoo, B., Ball, R., Shpanskaya, K., et~al.: Chexpert: A large chest radiograph dataset with uncertainty labels and expert comparison. In: Proceedings of the AAAI conference on artificial intelligence. vol.~33, pp. 590--597 (2019)

\bibitem{mimic}
Johnson, A.E., Pollard, T.J., Berkowitz, S.J., Greenbaum, N.R., Lungren, M.P., Deng, C.y., Mark, R.G., Horng, S.: Mimic-cxr, a de-identified publicly available database of chest radiographs with free-text reports. Scientific data  \textbf{6}(1), ~317 (2019)

\bibitem{sam}
Kirillov, A., Mintun, E., Ravi, N., Mao, H., Rolland, C., Gustafson, L., Xiao, T., Whitehead, S., Berg, A.C., Lo, W.Y., et~al.: Segment anything. In: Proceedings of the IEEE/CVF International Conference on Computer Vision. pp. 4015--4026 (2023)

\bibitem{doc2vec}
Le, Q., Mikolov, T.: Distributed representations of sentences and documents. In: International conference on machine learning. pp. 1188--1196. PMLR (2014)

\bibitem{li2024prototype}
Li, H., Song, J., Gao, L., Zhu, X., Shen, H.: Prototype-based aleatoric uncertainty quantification for cross-modal retrieval. Advances in Neural Information Processing Systems  \textbf{36} (2024)

\bibitem{blip}
Li, J., Li, D., Savarese, S., Hoi, S.: Blip-2: Bootstrapping language-image pre-training with frozen image encoders and large language models. In: International conference on machine learning. pp. 19730--19742. PMLR (2023)

\bibitem{li2020vispi}
Li, X., Cao, R., Zhu, D.: Vispi: Automatic visual perception and interpretation of chest x-rays. In: Medical Imaging with Deep Learning (2020)

\bibitem{sense}
Liu, B., Lu, Z., Wang, Y.: Towards medical vision-language contrastive pre-training via study-oriented semantic exploration. In: Proceedings of the 32nd ACM International Conference on Multimedia. pp. 4861--4870 (2024)

\bibitem{medsam}
Ma, J., He, Y., Li, F., Han, L., You, C., Wang, B.: Segment anything in medical images. Nature Communications  \textbf{15}(1), ~654 (2024)

\bibitem{radimagenet}
Mei, X., Liu, Z., Robson, P.M., Marinelli, B., Huang, M., Doshi, A., Jacobi, A., Cao, C., Link, K.E., Yang, T., et~al.: Radimagenet: an open radiologic deep learning research dataset for effective transfer learning. Radiology: Artificial Intelligence  \textbf{4}(5),  e210315 (2022)

\bibitem{lovt}
M{\"u}ller, P., Kaissis, G., Zou, C., Rueckert, D.: Joint learning of localized representations from medical images and reports. In: European Conference on Computer Vision. pp. 685--701. Springer (2022)

\bibitem{dino}
Oquab, M., Darcet, T., Moutakanni, T., Vo, H., Szafraniec, M., Khalidov, V., Fernandez, P., Haziza, D., Massa, F., El-Nouby, A., et~al.: Dinov2: Learning robust visual features without supervision. arXiv preprint arXiv:2304.07193  (2023)

\bibitem{radford2021learning}
Radford, A., Kim, J.W., Hallacy, C., Ramesh, A., Goh, G., Agarwal, S., Sastry, G., Askell, A., Mishkin, P., Clark, J., et~al.: Learning transferable visual models from natural language supervision. In: International conference on machine learning. pp. 8748--8763. PMLR (2021)

\bibitem{mura}
Rajpurkar, P., Irvin, J., Bagul, A., Ding, D., Duan, T., Mehta, H., Yang, B., Zhu, K., Laird, D., Ball, R.L., et~al.: Mura: Large dataset for abnormality detection in musculoskeletal radiographs. arXiv preprint arXiv:1712.06957  (2017)

\bibitem{nih}
Wang, X., Peng, Y., Lu, L., Lu, Z., Bagheri, M., Summers, R.M.: Chestx-ray8: Hospital-scale chest x-ray database and benchmarks on weakly-supervised classification and localization of common thorax diseases. In: Proceedings of the IEEE conference on computer vision and pattern recognition. pp. 2097--2106 (2017)

\bibitem{medclip}
Wang, Z., Wu, Z., Agarwal, D., Sun, J.: Medclip: Contrastive learning from unpaired medical images and text. arXiv preprint arXiv:2210.10163  (2022)

\bibitem{kl-cvr}
Wei, X., Vagena, Z., Kurtz, C., Cloppet, F.: Integrating expert knowledge with vision-language model for medical image retrieval. In: 2024 IEEE International Symposium on Biomedical Imaging (ISBI). pp.~1--4. IEEE (2024)

\bibitem{mcr}
Wei, Z., Jin, K., Zhou, X.: Masked contrastive reconstruction for cross-modal medical image-report retrieval. arXiv preprint arXiv:2312.15840  (2023)

\bibitem{yang2021uncertainty}
Yang, S., Fevens, T.: Uncertainty quantification and estimation in medical image classification. In: International conference on artificial neural networks. pp. 671--683. Springer (2021)

\bibitem{cxrclip}
You, K., Gu, J., Ham, J., Park, B., Kim, J., Hong, E.K., Baek, W., Roh, B.: Cxr-clip: Toward large scale chest x-ray language-image pre-training. In: International Conference on Medical Image Computing and Computer-Assisted Intervention. pp. 101--111. Springer (2023)

\bibitem{biomedclip}
Zhang, S., Xu, Y., Usuyama, N., Xu, H., Bagga, J., Tinn, R., Preston, S., Rao, R., Wei, M., Valluri, N., et~al.: Biomedclip: a multimodal biomedical foundation model pretrained from fifteen million scientific image-text pairs. arXiv preprint arXiv:2303.00915  (2023)

\bibitem{convirt}
Zhang, Y., Jiang, H., Miura, Y., Manning, C.D., Langlotz, C.P.: Contrastive learning of medical visual representations from paired images and text. In: Machine Learning for Healthcare Conference. pp. 2--25. PMLR (2022)

\bibitem{refers}
Zhou, H.Y., Chen, X., Zhang, Y., Luo, R., Wang, L., Yu, Y.: Generalized radiograph representation learning via cross-supervision between images and free-text radiology reports. Nature Machine Intelligence  \textbf{4}(1),  32--40 (2022)

\end{thebibliography}
\end{document}